\pdfoutput=1
\documentclass[11pt]{article}

\usepackage[]{acl}

\usepackage{times}
\usepackage{latexsym}
\usepackage{tablefootnote}
\usepackage[T1]{fontenc}
\usepackage[utf8]{inputenc}

\usepackage{microtype}
\usepackage{multirow}
\usepackage{wrapfig}
\usepackage{url}
\usepackage{latexsym}
\usepackage{color}
\usepackage{booktabs}
\usepackage{subfigure} 
\usepackage{multirow}
\usepackage{verbatim}
\usepackage{amsfonts}
\usepackage{float}
\usepackage{bm}
\usepackage{comment}
\usepackage{arydshln} 
\usepackage{array}
\usepackage{graphicx}  % DO NOT CHANGE THIS
\usepackage{amsmath}
\usepackage{blindtext}
\usepackage{wrapfig}
\usepackage{algorithmic}
\usepackage[lined,boxed,commentsnumbered,ruled,linesnumbered]{algorithm2e}
\definecolor{mygreen}{RGB}{102, 205, 170}
\definecolor{mypink}{RGB}{238, 162, 173}
\definecolor{myyellow}{RGB}{238, 232, 170}
\definecolor{myblue}{RGB}{164, 211, 238}

\title{MuGER$^2$: {M}ult{i}-{G}ranularity {E}vidence \textbf{R}etrieval and \textbf{R}easoning for Hybrid Question Answering }
\author{Yingyao Wang\textsuperscript{1}, Junwei Bao\textsuperscript{2}\thanks{~~Work was done during the first author’s internship at JD AI mentored by Junwei Bao: baojunwei001@gmail.com.}, Chaoqun Duan\textsuperscript{2}, Youzheng Wu\textsuperscript{2}, \\ \bf Xiaodong He\textsuperscript{2}, Tiejun Zhao\textsuperscript{1}\thanks{~~Corresponding author.}\\
    \textsuperscript{1}Harbin Institute of Technology \;\;  \textsuperscript{2}JD AI Research \\
    yywang@hit-mtlab.net \;\;  baojunwei001@gmail.com  \;\; tjzhao@hit.edu.cn
    }

\begin{document}
\maketitle

\begin{abstract}
Hybrid question answering (HQA) aims to answer questions over heterogeneous data, including tables and passages linked to table cells. The heterogeneous data can provide different granularity evidence to HQA models, e.t., column, row, cell, and link. Conventional HQA models usually retrieve coarse- or fine-grained evidence to reason the answer. 
Through comparison, we find that coarse-grained evidence is easier to retrieve but contributes less to the reasoner, while fine-grained evidence is the opposite.
To preserve the advantage and eliminate the disadvantage of different granularity evidence, 
we propose \textbf{MuGER$^2$}, a {\textbf{Mu}lti-\textbf{G}ranularity} \textbf{E}vidence \textbf{R}etrieval and \textbf{R}easoning approach.
In evidence retrieval, a unified retriever is designed to learn the multi-granularity evidence from the heterogeneous data. In answer reasoning, an evidence selector is proposed to navigate the fine-grained evidence for the answer reader based on the learned multi-granularity evidence. 
Experiment results on the HybridQA dataset show that {MuGER$^2$} significantly boosts the HQA performance.
Further ablation analysis verifies the effectiveness of both the retrieval and reasoning designs.\footnote{Our code is publicly available in \url{https://github.com/JD-AI-Research-NLP/MuGER2}.}.
\end{abstract}

%===========================================

%===========================================

%===========================================
%===========================================
% \begin{figure*}[t]
% \centering
% \subfigure[Table-G. ]{\includegraphics[width=0.195\textwidth]{fig/table-g.pdf}}
% \subfigure[Row-g. ]{\includegraphics[width=0.195\textwidth]{fig/row-g.pdf}}
% \subfigure[Cell-g. ]{\includegraphics[width=0.195\textwidth]{fig/cell-g.pdf}}
% \subfigure[Passage-g. ]{\includegraphics[width=0.195\textwidth]{fig/passage-g.pdf}}
% \subfigure[Multi-g. ]{\includegraphics[width=0.195\textwidth]{fig/multi-g.pdf}}
% \caption{.}
% \label{evidence}
% \end{figure*}
%===========================================

\section{Introduction}
\begin{figure}[t]
\setlength{\belowcaptionskip}{-0.3cm}
\centering
\includegraphics[width=0.48\textwidth]{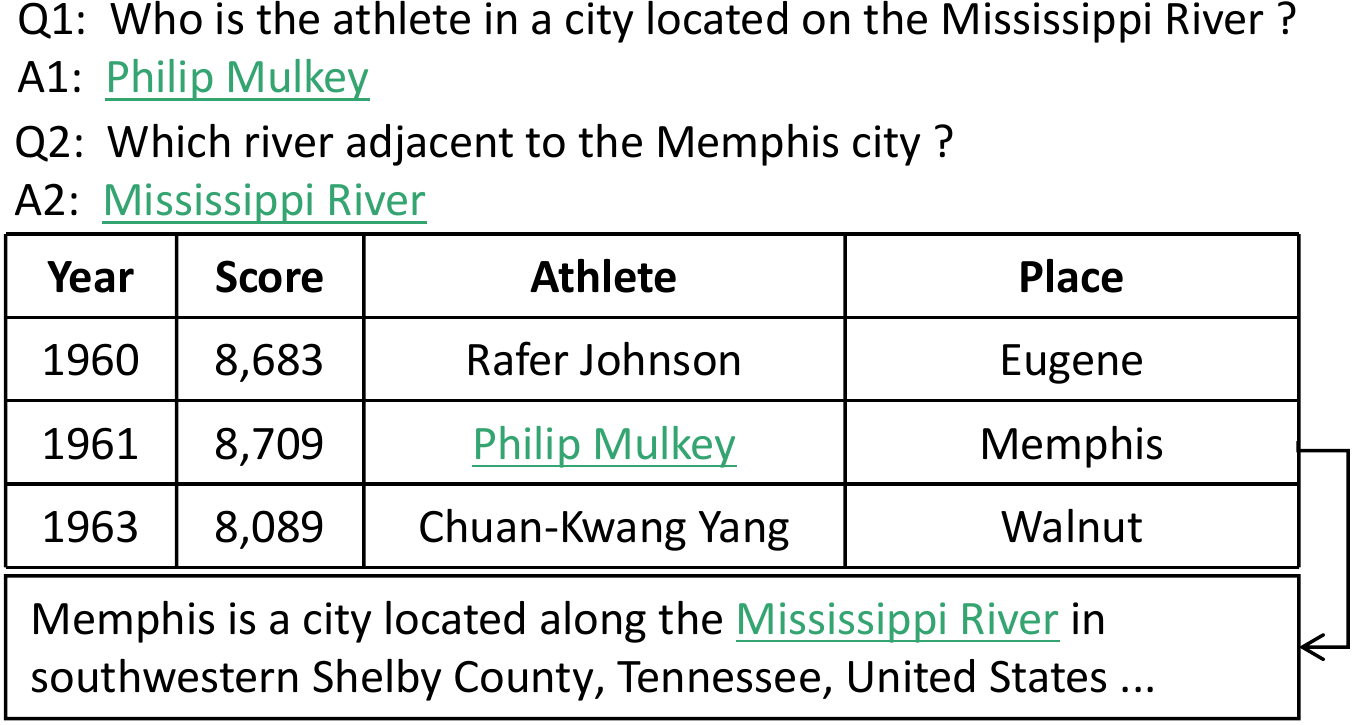}
\caption{Two example questions of HQA task.}
\label{task}
\end{figure}

%Traditional knowledge-based question answering systems derive answers to questions based on homogeneous knowledge, including structured knowledge bases or tables~\cite{yih2015semantic,jauhar2016tables,zhang2020table}, and unstructured text \cite{joshi2017triviaqa,dunn2017searchqa}, which fail to answer questions that require reasoning over heterogeneous data.
Traditional knowledge-based question answering systems derive answers to questions based on homogeneous knowledge, including knowledge graphs~\cite{bao2016constraint,gu2021beyond}, tables~\cite{jauhar2016tables,zhang2020table}, or passages~\cite{zhao-etal-2021-ror-read,zhou-etal-2022-opera} and achieve remarkable performances.
However, they neglect a more general scenario requiring reasoning over heterogeneous data to answer a question.
%To address this limitation, hybrid question answering (HQA) together with the HybridQA dataset is proposed \cite{chen-etal-2020-hybridqa}.
To investigate this challenge, \citet{chen-etal-2020-hybridqa} propose the hybrid question answering (HQA) and publish a corresponding dataset, HybridQA.
Specifically,  each question of HybridQA is aligned with a table and passages linked to table cells (abbreviated as links) as knowledge. The answer to the question may come from either table cells or links.
Figure~\ref{task} illustrates two examples of HybridQA. For the first one, the answer `\textit{Philip Mulkey}' comes from the cell, while for the second one, its answer `\textit{Mississippi River}' is derived from the link.

% \vspace{0.5cm}

%Conventional HQA methods usually first retrieve a \textit{mono-granularity} table region as evidence, e.g.,column,  row, cell, and link shown in Figure~\ref{motivation}, and then directly feed it into a reading comprehension (RC) model to reason the answer.
% \citet{zhang-etal-2020-table} first encode the whole table as evidence, and then answer questions based on its representation.
% \citet{glass2021capturing} 
%Hybrider \cite{chen-etal-2020-hybridqa} pinpoints a cell as evidence, and extracts answer from the cell content and its link with an RC model.
%Dochopper \cite{sun2021end} retrieves a text fragment as evidence, and extracts the answer with an RC model.
Existing  HQA models usually consist of two components, a \textbf{retriever} to learn evidence and a \textbf{reasoner} to leverage the evidence to derive the answer. Intuitively, the heterogeneous data shown in Figure~\ref{task} can provide different granularity evidence to the HQA models, such as coarse-grained column and row and fine-grained cell and link. 
Previous models always retrieve coarse or fine-grained evidence and directly use a span-based reading comprehension model to reason the answer.
For example, \citet{kumar2021multi} choose a coarse-grained region as the evidence, e.g., a table row.
On the contrary, \citet{chen-etal-2020-hybridqa} and \citet{sun2021end} focus on the fine-grained units,  table cell and link. 
Intuitively, compared with fine-grained evidence, coarse-grained evidence is easier to be accurately retrieved.
However, it contributes less to the reasoner's performance since it contains more noisy information.
The fine-grained evidence is just the opposite for the retriever and the reasoner.

\begin{figure}[t]
\setlength{\belowcaptionskip}{-0.4cm}
\centering
\includegraphics[width=0.45\textwidth]{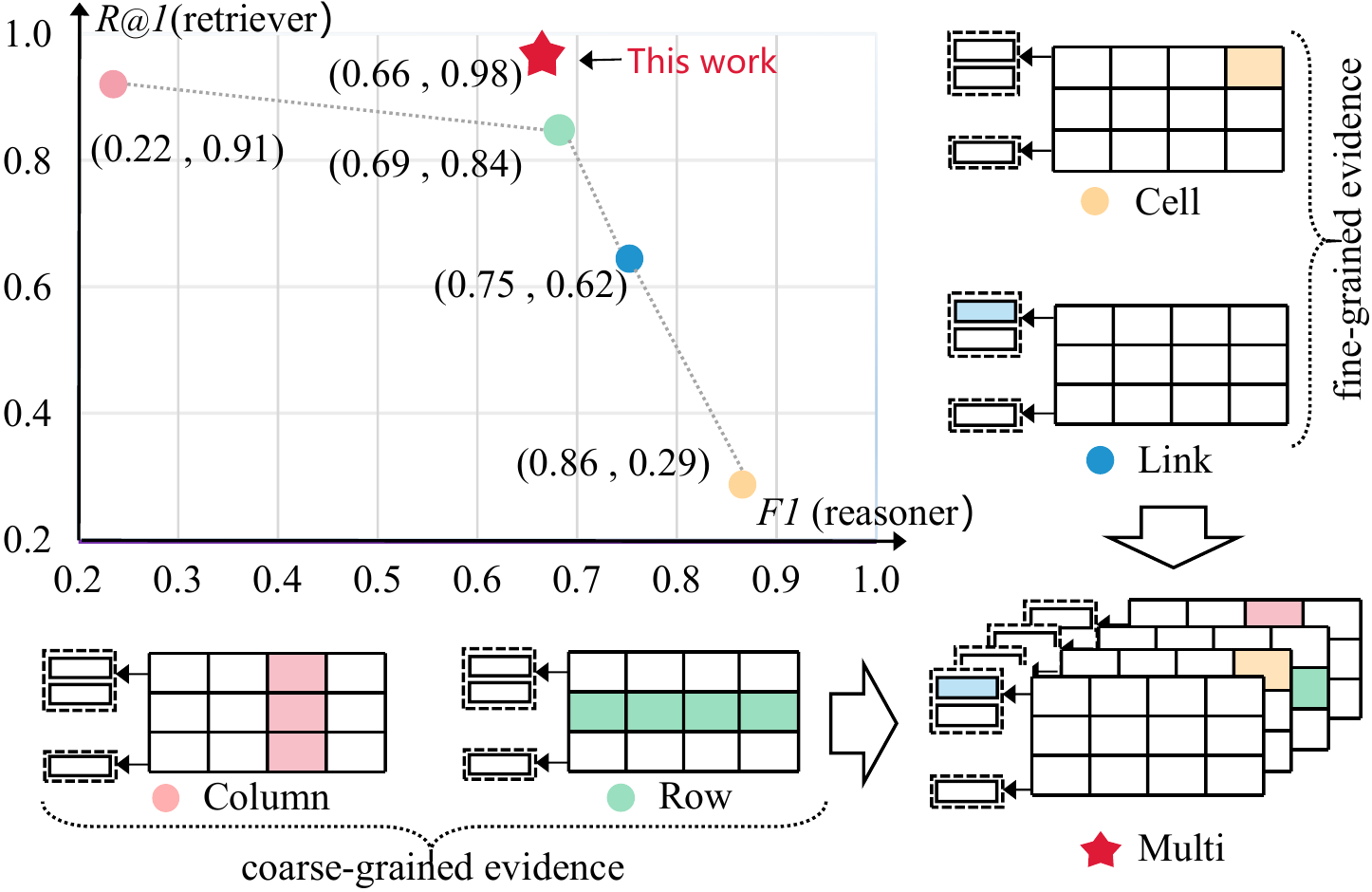}
\caption{The retriever and reasoner performance with different granularity evidence.
% The answer recall $R@1$ is defined to evaluate the performance of the retriever, which denotes the probability that the retrieved evidence contains the answer. 
 Note that the $F1$ score in this figure is calculated based on the \textbf{oracle} evidence to evaluate the answer reasoning performance. 
Intuitively, $R@1 \times F1$ reflects the end-to-end HQA performance. }
\label{motivation}  
\end{figure}

To prove this argument, we conduct experiments on the development set of HybridQA to analyze the retriever and the reasoner performance\footnote{In these experiments, the retriever and reader models are all initialized with BERT-base.} with different granularity evidence.  The results of evidence retrieval and answer reasoning are evaluated with the retrieval recall (R@1) and the F1 score, respectively.
As shown in Figure~\ref{motivation}, the coarse-grained evidence achieves better retrieval recalls but lower F1 scores than the fine-grained evidence. With the granularity changing from coarse to fine, the  F1 score increases while the R@1 declines.  
The results are consistent with our hypothesis.

To preserve the advantage and eliminate the disadvantage of different granularity evidence, we propose \textbf{MuGER$^2$}, a {\textbf{Mu}lti-\textbf{G}ranularity} \textbf{E}vidence \textbf{R}etrieval and \textbf{R}easoning approach for HQA.
In the retrieval stage, a unified retriever is designed to learn the multi-granularity evidence from the heterogeneous data involving columns, rows, cells, and links. Compared with existing methods, the multi-granularity evidence preserves the retrieval recall and provides more aspects of information to reason the answer. In the reasoning stage, to avoid the redundant information in the multi-granularity evidence that lowers the answer reader's performance, an evidence selector (E-SEL) is designed to navigate the fine-grained evidence for the reader by fusing the learned multi-granularity evidence. Together with our multi-granularity evidence retrieval and reasoning designs, our MuGER$^2$ preserves both the retrieval recall and the F1 score and further boosts the end-to-end HQA performance.

We conduct extensive experiments on the HybridQA dataset to verify the effectiveness of our proposed MuGER$^2$.
Experiment results show that MuGER$^2$ achieves remarkable improvements, which outperforms a publicly available strong baseline by $10.0\%$ EM and $12.9\%$ F1 scores on the end-to-end HQA performance.
Ablation studies verify the effectiveness of the proposed multi-granularity evidence retrieval and the evidence selector.
Moreover, results compared with different mono-granularity baselines prove the effectiveness of multi-granularity evidence for HQA.

Our contributions are as follows:
(1) We analyze the limitation of HQA systems which solely rely on only one certain granularity evidence in both the retriever and reasoner and propose a {{mu}lti-{g}ranularity} {e}vidence {r}etrieval and {r}easoning approach, which boosts the end-to-end HQA results.
(2) We propose a joint retrieval method that improves the evidence retrieval performance and an evidence selector to accurately navigate the fine-grained evidence from the multi-granularity information to preserve the reader's performance.
(3) We conduct extensive experiments on the HybridQA dataset and show the effectiveness of {MuGER$^2$}.  

% \begin{itemize}
%   \item We formalize HQA as evidence retrieval and answer reasoning, and find the issue of mono-granularity methods via empirically analyzing the trade-off between the two stages.
%   \item We propose {MuGER$^2$},
%   a {{mu}lti-{g}ranularity} {e}vidence {r}etrieval and {r}easoning approach, 
%   which upgrades mono- to multi- granularity evidence as bridge from question to answer.
%   \item We conduct extensive experiments on HybridQA dataset and show the effectiveness of {MuGER$^2$}. We will release our code to further the research on HQA for the community. 
% \end{itemize}

%===========================================
\begin{figure*}[t]
\setlength{\belowcaptionskip}{-0.3cm}
\centering
\includegraphics[width=0.99\textwidth]{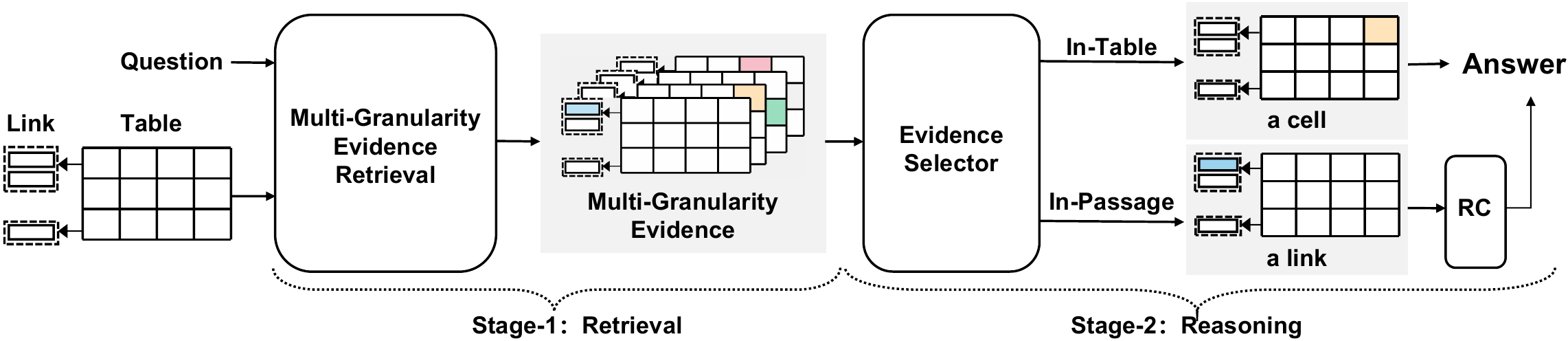}
\caption{The framework of MuGER$^2$ which performs multi-granularity evidence retrieval and answer reasoning. Evidence of four kinds of granularity including $\{ \colorbox{mypink}{\texttt{column}}, \colorbox{mygreen}{\texttt{row}}, \colorbox{myyellow}{\texttt{cell}}, \colorbox{myblue}{\texttt{link}}\}$ are in color. } 
\label{minda}  
% \vspace{-2em}
\end{figure*}
%===========================================

\section{Methodology}
\subsection{Task: Hybrid Question Answering}
The hybrid question answering (HQA) aims to tackle the answer reasoning over heterogeneous data, as shown in Figure~\ref{task}.
The input of this task includes a question $\mathcal{Q}$, a table $\mathcal{T}$, and a set of links $\mathcal{L}$.
Specifically, the table $\mathcal{T}$ consists of cells $\{c_{i,j}\}_{i=1,j=1}^{i=M,j=N}$, where $M$ and $N$ are the numbers of rows and columns. Each column has a header $h_j$ to describe the field of the cells $\{c_{i,j}\}_{i=1}^{i=M}$. 
A table cell $c_{i,j}$ may link to a subset of links $\mathcal{L}_{i,j} \subseteq \mathcal{L}$. 
Given the input, an HQA system aims to return the answer text $\mathcal{A}$ of the question $\mathcal{Q}$ by reasoning on the heterogeneous information. In this work,  $\mathtt{In}\text{-}\mathtt{Table}$ and $\mathtt{In}\text{-}\mathtt{Passage}$ respectively represent that the answer is in a cell or in a link.

\subsection{Model Overview}
\label{overview}
As described above, coarse-grained evidence is easier to be derived by the retriever and it makes less contribution to the reasoner while the fine-grained is just the opposite.
To preserve their advantage and get rid of their disadvantage, we propose {MuGER$^2$}, a {{mu}lti-{g}ranularity} {e}vidence {r}etrieval and reasoning approach.
As shown in Figure~\ref{minda}, in the retrieval stage, MuGER$^2$ adopts a unified retriever to learn the multi-granularity evidence.
In the reasoning stage, an evidence selector is designed to navigate the fine-grained evidence for the reader based on the multi-granularity evidence.
The details of these modules are introduced in the following sections.

\subsection{Stage-1: Multi-Granularity Evidence Retrieval }
In this section, we first introduce the definition of the multi-granularity evidence and then introduce the retrieval model of MuGER$^2$.
% Firstly, we setup a separate retrieval (SR) method as the baseline. Then we introduce the joint retrieval (JR) method used in our MuGER$^2$, which further benefit from the contrastive learning (CL). 
% The retrievers in this stage are based on the pre-trained BERT~\cite{devlin2018bert} encoder.
% We propose multi-granularity (multi-g) evidence retrieval which decomposes retrieval procedure into multiple retrievers, which respectively retrieve atomic evidence of different granularity.
\subsubsection{ Multi-Granularity Evidence\label{concept_section}}
\label{evi}
% We propose an answer reasoning framework to address this task, and the algorithm is given in Alg~\ref{al1}. 
According to the characteristics of the HQA,
we define the multi-granularity evidence as $\mathcal{E}$. $\mathcal{E}$ involves four kinds of evidence with different granularity which is denoted as
$\mathcal{E}$$=$$\langle \mathcal{E}_{\mathtt{col}}$, $\mathcal{E}_{\mathtt{row}}$, $ \mathcal{E}_{\mathtt{cell}}$,   $\mathcal{E}_{\mathtt{link}}\rangle$.
% $\mathcal{E}$$=$$\langle \mathcal{E}_{\mathtt{column}}$, $\mathcal{E}_{\mathtt{row}}$, $ \mathcal{E}_{\mathtt{cell(value)}}$,  $\mathcal{E}_{\mathtt{cell(link)}}$,  $\mathcal{E}_{\mathtt{link}}\rangle$.
%These evidence belong to four levels of granularity.
% The multi-granularity evidence $\mathcal{E}$ is defined  as $\mathcal{E}=
% \langle \mathcal{E}_{\mathtt{column}},\mathcal{E}_{\mathtt{row}},  \mathcal{E}_{\mathtt{cell(value)}}, \mathcal{E}_{\mathtt{cell(link)}}, \mathcal{E}_{\mathtt{link}}\rangle$, which contains five types of evidence.
%  Note that, the five types of evidence belong to four kinds of granularity as shown in Figure~\ref{minda}.
Specifically, $\mathcal{E}_{\mathtt{col}}$ indicates the {column} which includes the answer; 
$\mathcal{E}_{\mathtt{row}}$ represents the {row} that contains the answer; 
The evidence $\mathcal{E}_{\mathtt{cell}}$ means the cell that contains the answer in its cell value; and $\mathcal{E}_{\mathtt{link}}$ indicates the {link}  containing the answer.
\subsubsection{Evidence Retriever}
% % Since our model leverages five kinds of evidence, the retrieval stage can be formulated as:
% To integrate the above mentioned five kinds of evidence into our method, we adopt their joint probabilities to model the evidence retriever and formulate it as follows:
% % we adopt the joint probability distribution over them to model the evidence retriever and formulate it as follows:
To integrate the above mentioned four kinds of evidence into our method, we adopt their joint probabilities to model the evidence retriever.
Formally, given a question $\mathcal{Q}$ and the multi-granularity evidence candidate $\mathcal{E}$ derived from $\mathcal{T}$ and $\mathcal{L}$, the evidence retriever can be formulated as follows:
\begin{equation}
p_{ret}(\mathcal{E}|\mathcal{Q},\mathcal{T},\mathcal{L})=\prod_{t}p_{e}(\mathcal{E}_{t}|\mathcal{Q},\mathcal{T},\mathcal{L})
\end{equation}
where $t\in\{\mathtt{col}, \mathtt{row}, \mathtt{cell}, \mathtt{link}\}$, $\mathcal{E}_{t}$ denotes $t$-kind element of $\mathcal{E}$, and  $p_{e}(\mathcal{E}_{t}|\mathcal{Q},\mathcal{T},\mathcal{L})$ represents the probability that $\mathcal{E}_{t}$ is the evidence.
% of the candidate evidence $\mathcal{E}_{t}$ containing the answer to the question $\mathcal{Q}$.
To compute $p_{e}(\mathcal{E}_{t}|\mathcal{Q},\mathcal{T},\mathcal{L})$, we further introduce a retrieval score $s_{t}$ for $\mathcal{E}_{t}$, which weighs the possibility that $\mathcal{E}_{t}$ contains the answer, and adopt it as follows:
\begin{equation}
    p_{e}(\mathcal{E}_{t}|\mathcal{Q},\mathcal{T},\mathcal{L})=s_{t}/Z_{t}
\end{equation}
where $Z_{t}$ is a normalization factor and equal to the sum of retrieval scores of all $t$-kind evidence.
Given that for different evidence $Z_{t}$ is consistent, we omit it in our implementation, and only use the retrieval score for evidence selection.
% For simplicity, we substitute $p_{e}(\mathcal{E}_{t})$ when referring to $p_{e}(\mathcal{E}_{t}|\mathcal{Q},\mathcal{T},\mathcal{L})$ in the following.
%$\mathbf{h}_t \in \mathbb{R}^{K}$ indicates the representation of the candidate evidence, and $\mathbf{W}_t \in \mathbb{R}^{K}$ are trainable parameters.

To compute all retrieval scores for the multi-granularity evidence $\mathcal{E}$, a straightforward method is to train four separate models and predict the different granularity evidence respectively.
However, this method neglects relations among different granularity evidence.
To tackle this challenge, we propose to adopt a unified framework for all kinds of evidence and train it in a joint way.

Specifically, we first utilize BERT~\cite{devlin2018bert} as the encoder to learn the representation for each evidence.
Formally, given evidence $\mathcal{E}_{t}$, we concatenate $t$, $\mathcal{Q}$, and $\mathcal{E}_{t}$ with ``[SEP]'' into a sequence and prepend ``[CLS]'' at its start.
Subsequently, we feed the sequence into BERT.
Finally, we adopt the max-pooling operation over all tokens of the BERT output to obtain the sequence representation $\mathbf{h}_{t}$.
This process is denoted as follows:
% we first utilize the BERT~\cite{devlin2018bert} encoder followed by the max-pooling operation to learn the representation $\mathbf{h}_{t}$ of the candidate evidence $\mathcal{E}_{t}$:
\begin{equation}
\begin{split}
    \mathcal{S}_{t}=[\text{[CLS]}&;t;\text{[SEP]};\mathcal{Q};\text{[SEP]};\mathcal{E}_{t}] \\
    \mathbf{h}_{t} =&  \mathtt{MaxP}(\mathtt{BERT}(\mathcal{S}_{t}))
\end{split}
%  \mathbf{h}_{t} =  \mathtt{MaxP}(\mathtt{BERT}([\text{[CLS]};t;\text{[SEP]};\mathcal{Q};\text{[SEP]};\mathcal{E}_{t}]))   
\end{equation}
where $t$ is inserted into the input sequence to indicate the kind of the evidence. After obtaining the representation of $\mathcal{E}_{t}$, we feed it into a linear projection followed by the sigmoid function to compute the retrieval score as follows, where $\mathbf{W}\in \mathbb{R}^{K}$ is a $K$-dimensional  trainable parameter.
% Each of the retrievers first learns the representation $\mathbf{h}_{t}$ of the candidate evidence $\mathcal{E}_{t}$, and then calculate the binary cross entropy loss as the training objective. Each retriever is formulated as
\begin{equation}
% p_{e}(\mathcal{E}_{t})\propto(\mathbf{h}_{t}^{\top}\mathbf{W}_t)
s_{t}=\mathtt{sigmoid}{(\mathbf{h}_{t}^{\top}\mathbf{W})}
%\mathbf{h}_{t} =  \mathtt{MaxP}(\mathtt{BERT}([\text{[CLS]};\mathcal{Q};\text{[SEP]};\mathcal{E}_{t}]))
\label{sr}
\end{equation}

\subsubsection{Content of Different Evidence}
For $\mathcal{E}_{t}$, it refers to different content when $t$ denotes different kinds of evidence as follows. Notably, the granularity of certain evidence is determined by what kinds of candidates are ranked, e.g., cells or rows, but NOT what content a candidate uses. 
% The contents of $\mathcal{E}_{t}$ of different granularity are defined as follows:
% \paragraph{Column}

\paragraph{Column}
Column evidence aims to take the meaning of a specific column header in the table into account. The column header usually describes the fields of an answer.
Therefore, we utilize the table header as column evidence.
Namely, $\mathcal{E}_\mathtt{col}\!=\!h_j$.
% The column retrieval aims to select a suitable table header, thus we define the $\mathcal{E}_\mathtt{col} = h_j$.

\paragraph{Link}
Link evidence indicates a passage that a table cell links to.
It is another form of knowledge in the HQA task and we assign each $\mathcal{E}_\mathtt{link}$ candidate a specific passage.

\paragraph{Cell} 
We formalize cell evidence $\mathcal{E}_\mathtt{cell}$ as a concatenation of its header $h_j$, its neighbors in the same row $\{h_j,c_{i,j}\}_{j=1}^{j=N}$, and its links $\mathcal{L}_{i,j}$. Namely, $\mathcal{E}_\mathtt{cell}\!=\![h_j;$ $\text{[SEP]};$ $\{h_j,c_{i,j}\}_{j=1}^{j=N};$ $\text{[SEP]};$ $\mathcal{L}_{i,j}]$, where ``[SEP]'' is the separator. In cell encoding, the above neighbor information refers to the values of other cells in the same row without their linked passages. Neighbor information and the corresponding header of a cell are used as its positional information (row, column) which is essential to distinguish cells, especially for those with the same value but in different positions. 
This process is also applied to the modeling of row evidence.

\paragraph{Row} Traditional methods usually concatenate all row cells into an input sequence to score a table row evidence. However, in the HQA task, a table row may link to several passages, and directly concatenating all these contents will result in an overlength of input sequences. Therefore, we score a row based on its cells. Therefore, to encode and score a row candidate, we first encode all its cells separately and then aggregate the cell scores by maximization operation to obtain the row score.
Namely, $\mathcal{E}_\mathtt{row}$ includes a set of  $\mathcal{E}_\mathtt{cell}$.

% where $\mathbf{W}\in \mathbb{R}^{K}$ means our model use a shared parameter to learn the five kinds of evidence.
%Denote $y_t \in \{0,1\}$ as the label of the candidate evidence, the training loss is calculated as follows:

%Note that, since only the answers are annotated in the HybridQA dataset, the evidence labels are obtained by distant supervision method like \citet{chen-etal-2020-hybridqa} which basically follows the principle that the positive evidence contains the answer.

\subsubsection{Training}
\paragraph{Joint Training for Multi-Granularity}
As mentioned above, we only adopt the retrieval score for the evidence selection.
Therefore, the core of training of evidence retriever is to learn a model to produce the retrieval score.
Since the retrieval score $s_{t}$ means the probability that the evidence $\mathcal{E}_{t}$ contains the answer, we treat the retrieval score prediction as a binary classification and use the binary cross entropy (BCE) loss as the training objective to learn the model as follows:
% To train the unified retriever, an intuitive method is to treat the five retrieval tasks as binary classification and use the binary cross entropy (BCE) loss as the training objective.
% For each kind of evidence, the probability $p_{e}(\mathcal{E}_{t})$ is first obtained from the retriever, then the binary cross entropy loss of $\mathcal{E}_{t}$ is  calculate as follows:
\begin{equation}
    \mathtt{Loss}_t^{bce}= {\mathtt{BCE}(s_{t}, y_{t})} 
    \label{all_loss}
\end{equation}
where $y_t \in \{0,1\}$ is the label of $\mathcal{E}_{t}$ indicating whether the evidence $\mathcal{E}_{t}$ contains the answer. Note that, since only the answer texts are annotated in the HybridQA dataset, we obtain the evidence labels by distant supervision method like \citet{chen-etal-2020-hybridqa} which basically follows the principle that the positive evidence contains the answer.

Furthermore, to consider the relative information among different granularity evidence, we propose to leverage a unified retriever to model them and learn it in a joint way:
% Therefore, the final objective function can be formulated as:
% the unified retriever learns the five retrieval tasks in a joint way. Therefore, the training objective of the unified retriever can be formulated as 
\begin{equation}
    \mathtt{Loss}^{bce}= \sum_{t}{\mathtt{Loss}^{bce}_t} 
    \label{bin}
\end{equation}

\paragraph{Contrastive Learning within Each Granularity}
The above joint training method is capable of capturing the relation among different granularity evidence.
However, it neglects the relations among different candidate evidence within the same granularity.
Because it depends on the BCE loss, which regards each candidate evidence as an individual instance and does not require any involvement of other instances besides the given evidence.
% Although the above joint training way can capture the relations of cross-granularity evidence, the binary classification loss still regards each item of evidence candidates as individuals, which misses the relations of inner-granularity evidence.
% To alleviate this problem, we propose to leverage contrastive learning (CL) ~\cite{1640964} among inner-granularity evidence candidates to model their relative similarities and further enhance the model's representation and scoring capability.
To alleviate this problem, we propose to leverage contrastive learning (CL) ~\cite{1640964} within candidate evidence with the same granularity to model their relative similarities and further enhance the model's representation and scoring capability.

To construct the training objective, we first define a function $\mathtt{sim}(\cdot)$ to denote the similarity between different evidence as follows:
\begin{align}
    \mathtt{sim}({\mathbf{h}_t},{\mathbf{h}_{t}^{'}})= \mathbf{h}_t \mathbf{W} + \mathbf{h}_{t}^{'}\mathbf{W}
\end{align}
% where $\mathbf{W}$ is a trainable parameter.

It also requires building positive and negative instances for a given evidence candidate.
For positive instance, following ~\citet{gao2021simcse}, we set $\mathcal{E}_{t}^{+}\!=\!\mathcal{E}_{t}$.
Since there are dropout masks placed on fully-connected layers as well as attention probabilities during training of BERT,
feeding the same input to the encoder twice we get two different representations $\mathbf{h}_t$ and $\mathbf{h}_t^{+}$ for it.
To obtain the negative samples $\mathcal{E}_{t}^{-}$ for the given evidence, like most existing work with contrastive learning~\cite{gao2021simcse}, we regard other evidence in the same batch as negative ones and the representation of $\mathcal{E}_{t}^{-}$ is denoted as $\mathbf{h}_t^{-}$.

After that, we take a cross-entropy objective with in-batch negatives.
Formally, given a batch, we split it into four groups and each one corresponds to a granularity. 
Let $D$ denote all instances in a group, the training objective is:
\begin{align}
    \label{loss2}
    \mathtt{Loss}_{t}^{cl}=- \mathtt{log}\frac{e^{\mathtt{sim}(\mathbf{h}_t ,\mathbf{h}_t^{+})/ \tau }}{\sum_{\mathbf{h}_t^{-} \in D}e^{\mathtt{sim}(\mathbf{h}_t,\mathbf{h}_t^{-})/ \tau } }
\end{align}
where $\tau$ is a temperature hyperparameter.
Actually, we compute four loss as Eq.~(\ref{loss2}) based on four groups and jointly learn them as follows:
\begin{align}
    \mathtt{Loss}^{cl}=\sum_{t}\mathtt{Loss}_{t}^{cl}
\end{align}
Finally, we combine $\mathtt{Loss}^{bce}$ and $\mathtt{Loss}^{cl}$ to produce the final objective function:
\begin{align}
    \mathtt{Loss}=\mathtt{Loss}^{bce}+\mathtt{Loss}^{cl}
\end{align}

\subsection{Stage-2: Answer Reasoning with the Multi-Granularity Evidence}
As shown in Figure~\ref{minda}, answer reasoning module $p_{rea}(\mathcal{A}|\mathcal{E})$ is responsible for deriving answer based on the output of the evidence retriever.
This module consists of two essential components, evidence selector and reader.
Algorithm~\ref{al1} shows the whole workflow of this module.
% Algorithm~\ref{al1} shows the answer reasoning procedure $p_{rea}(\mathcal{A}|\mathcal{E})$ which aims to leverage the multi-granularity evidence distribution $p_{ret}(\mathcal{E}|\mathcal{Q},\mathcal{T},\mathcal{L})$ (line 1) to derive natural language answer $\mathcal{A}$.  We introduce the two components of the reasoning stage, E-SEL and RC, in the following sections.

\subsubsection{Evidence Selector\label{ATR}}
% Although the multi-granularity evidence achieves a high retrieval recall, its redundant information causes weak RC performance.

As analysed above, multi-granularity evidence is easier to be accurately scored by the retriever.
However, they contain too much redundant information and brings burden to the reader.
% To alleviate this issue, the evidence selector (E-SEL) is proposed to fuse the multi-granularity evidence.
To alleviate this issue, the evidence selector (E-SEL) is proposed to navigate the fine-grained evidence for the reader based on the multi-granularity evidence.

% Specifically, we assign each cell two scores,  $s_{tab}^{i,j}$ and $s_{pass}^{i,j}$, to respectively indicate the probability that the cell value and the cell links contain the answer.

Specifically, we define two scores  $s_{tab}^{i,j}$ and $s_{pass}^{i,j}$ to respectively indicate the probabilities that the answer is in the cell $c_{i,j}$ and the links $\mathcal{L}_{i,j}$.
% These two scores are calculated by considering multiple granularity information, including the cell itself and the row and column in which the cell is located in. 
When calculating the scores, we take the multi-granularity evidence into account. Suppose that the link set $\mathcal{L}_{i,j}$ has $X$ links denoted as $\{l_x, x\in[1,X]\} $. 
The $s_{tab}^{i,j}$ and $s_{pass}^{i,j}$ are calculated as follows:
%Specifically, evidence retriever assigns each evidence candidate a retrieval score.
%Namely,for each table cell $c_{i,j}$, it is attached with several scores, including $s_{col}^{j}$, $s_{row}^{i}$, $s_{cval}^{i,j}$, $s_{clink}^{i,j}$, etc.
%%They can be regarded as weights from different aspects to measure the possibility that $c_{i,j}$ contains the answer.
%We adopt some of them to compute two scores for $c_{i,j}$ as follows:
\begin{align}
    \begin{split}
        s_{tab}^{i,j} &= s_{\mathtt{col}}^{j}+s_{\mathtt{row}}^{i}+s_{\mathtt{cell}}^{i,j} \\
        s_{pass}^{i,j} &= s_{\mathtt{col}}^{j}+s_{\mathtt{row}}^{i}+\mathtt{Max}_{x=1}^{x=X}{(s_{\mathtt{link}}^{x})}
    \end{split}
    \label{esel}
\end{align}
where $s_{col}^{j}$, $s_{row}^{i}$, $s_{cell}^{i,j}$, and $s_{link}^{x}$ are retrieval scores.
% After that, we further obtain two candidate target cells with the global maximum scores $s_{tab}$ and $s_{pass}$:
% After that, E-SEL is able to judge the answer type based on these scores as follows:
After that, we can obtain two global maximum scores $s_{tab}$ and $s_{pass}$ as follows:
\begin{align}
\begin{split}
    s_{tab}&=\mathtt{Max}_{i=1,j=1}^{i=M,j=N}(s_{tab}^{i,j}) \\
    s_{pass}&=\mathtt{Max}_{i=1,j=1}^{i=M,j=N}(s_{pass}^{i,j})
\end{split}
\end{align}
%We regards these two scores as probabilities that the answer is within the table and passages.
These two scores, $s_{tab}$ and $s_{pass}$, indicate that the answer is in the table or passages by simply comparing their values.
Meanwhile, the cell and link corresponding to them are considered as the navigated evidence for the reader.
Based on $s_{tab}$ and $s_{pass}$, E-SEL is able to judge the answer type of the current question as follows:
\begin{equation}
\mathtt{E\text{-}SEL}(p_{ret}(\cdot))=
\left\{
\begin{aligned}
& \mathtt{In}\text{-}\mathtt{Table}, & s_{tab} > s_{pass} \\
& \mathtt{In}\text{-}\mathtt{Passage}, & s_{tab} \leq s_{pass}
\end{aligned}
\right.
\end{equation}
where $\mathtt{In}\text{-}\mathtt{Table}$ denotes that the answer to the question is in the cell, while $\mathtt{In}\text{-}\mathtt{Passage}$ indicates that the answer is in the link.

% , i.e., $\mathtt{In}\text{-}\mathtt{Table}$ and $\mathtt{In}\text{-}\mathtt{Passage}$.
% To this end, we first compute two scores for cell $c_{i,j}$ as follows:

% Afterwards, the fine-grained cell or link is selected based on the answer type and the scores. If the answer is in $\mathtt{In}\text{-}\mathtt{Table}$, the cell value of $c^1$ with score $s^1$ is returned as the answer, otherwise (line 6-10) the top-1 link of cell $c^2$ with score $s^2$ is selected to derive answer by the RC model.
After obtaining the answer type, we can navigate the cell or link as the final fine-grained evidence to derive the answer.
Formally, if the type is $\mathtt{In}\text{-}\mathtt{Table}$, the cell with score $s_{tab}$ will be selected and its value will be returned as the answer.
Otherwise, the link with the score $s_{pass}$  will be selected and it will be feed to the reader to extract answer.

\subsubsection{Reader}
% \subsubsection{Reading Comprehension}
% Following previous work~\cite{chen-etal-2020-hybridqa,sun2021end}, 
% We adopt BERT to train the span-based reading comprehension (RC) models like ~\cite{chen-etal-2020-hybridqa,sun2021end}.
% An RC model $\mathtt{RC}(\mathcal{Q},{p})$ derives answer by extracting a span from the retrieved link $p$ as answer based on the given question $\mathcal{Q}$ (line 9).
For the reader $\mathtt{RC}(\mathcal{Q},{l})$, following existing works~\cite{chen-etal-2020-hybridqa,sun2021end}, we implement it with a span-based reading comprehension model which adopt BERT as backbone.

\begin{algorithm}[t]
\caption{Answer Reasoning}
\label{al1}
\SetKwData{Index}{Index}
% \LinesNumbered 
\KwIn{question $\mathcal{Q}$, table $\mathcal{T}$, 
links $\mathcal{L}$}
\KwOut{answer text $\mathcal{A}$}
calculate $p_{ret}(\mathcal{E}|\mathcal{Q},\mathcal{T},\mathcal{L})$\;
\If {$\mathtt{E\text{-}SEL}(p_{ret}(\cdot))$==$\mathtt{In}\text{-}\mathtt{Table}$}
{
    get the cell $c_{tab}$ with score $s_{tab}$\;
    return cell value of $c_{tab}$ as answer $\mathcal{A}$
}
\ElseIf {$\mathtt{E\text{-}SEL}(p_{ret}(\cdot))$==$\mathtt{In}\text{-}\mathtt{Passage}$}
{
    get the link $l_{pass}$ with score $s_{pass}$\;
    % $l$ top-1 link of $c_{i,j}$ based on $p_{ret}(\cdot)$\;
   % $l$ is the passage which $c_{i,j}$ links to\;
    return $\mathtt{RC}(\mathcal{Q},l)$ as answer $\mathcal{A}$
}
\end{algorithm}

\begin{table*}[ht]
\setlength{\belowcaptionskip}{-0.4cm}
	\centering
	\small
	\linespread{1.5}
% 	\resizebox{0.99 \textwidth}{!}{
%     \setlength{\tabcolsep}{1mm}{
	\begin{tabular}{l|ccc|ccc}
		\toprule
		\toprule
		\specialrule{0em}{1pt}{1pt}
		\multirow{3}{*}{\bf Model}     & \multicolumn{3}{c|}{\bf Dev}      & \multicolumn{3}{c}{\bf Test}            \\ 
		\specialrule{0em}{1pt}{1pt}
		\cline{2-7} 
	     \specialrule{0em}{1pt}{1pt}
	     & In-Table & In-Passage & Total      & In-Table & In-Passage & Total  \\ 
	     \specialrule{0em}{1pt}{1pt}
	     \cline{2-7}
	     \specialrule{0em}{1pt}{1pt}
	     & EM / F1 & EM / F1 & EM / F1            & EM / F1 & EM / F1 & EM / F1      \\ 
	     \specialrule{0em}{1pt}{1pt}
	     \midrule
	     \specialrule{0em}{1pt}{1pt}
	     MQA-QG (unsupervised) &-- / --& -- / -- & -- / -- &36.2 / 40.6& 19.8 / 25.0 & 25.7 / 30.5 \\
	     \specialrule{0em}{1pt}{1pt}
	     \midrule
	     \specialrule{0em}{1pt}{1pt}
	     Table-Only&  14.7 / 19.1 & 2.4 / 4.5 & 8.4 / 12.1 & 14.2 / 18.8 & 2.6 / 4.7 & 8.3 / 11.7\\ 
	     Passage-Only& 9.2 / 13.5 & 26.1 / 32.4 & 19.5 / 25.1 & 8.9 / 13.8 & 25.5 / 32.0 & 19.1 / 25.0 \\ 
	%     BM25 + ETC&  -- / -- & -- / -- & 24.8 / 29.1 &-- / --& -- / --& 25.5 / 31.1 \\ 
     %	 Dense + ETC&  -- / --& -- / --& 37.0 / 43.5 &-- / --& -- / --& 34.1 / 40.3 \\ 
     %	 Sequential (ETC)&  -- / --& -- / --& 39.4 / 44.8 &-- / --& -- / --& 37.0 / 43.0\\ 
     	 Hybrider (BERT-base, $\tau$=0.9)& 51.5 / 58.6 & 40.5 / 47.9 & 43.7 / 50.9 &52.1 / 59.3 &38.1 / 46.3 & 42.5 / 50.2\\ 
     	 Hybrider (BERT-large, $\tau$=0.8)& 54.3 / 61.4 & 39.1 / 45.7  & 44.0 / 50.7 &56.2 / 63.3 &37.5 / 44.4 & 43.8 / 50.6\\
     	 Dochopper & -- / --& -- / -- & 47.7 / 55.0 & -- / -- &-- / --& 46.3 / 53.3\\
      
%      	 MATE  & 68.6 / 74.2 & 62.8 / 71.9 & \bf 63.4 / \bf 71.0 & 66.9 / 72.3& 62.8 / 71.9 &\bf 62.8 / \bf 70.2 \\ 
         \specialrule{0em}{1pt}{1pt}
         \midrule
         \specialrule{0em}{1pt}{1pt}
     	
     	% Ours (top1hop w/o pred00) & 54.5 / 61.9 & 48.6 / 58.7 & 49.6 / 58.5 &-- / -- &-- / -- & -- / --\\ 
     	% Ours (top2hop w/o pred00) & 54.4 / 61.9 & 50.6 / 61.0 & 50.8 / 59.9 &-- / -- &-- / --& -- / --\\
     	% Ours (top3hop w/o pred00) & 54.7 / 62.2 & 51.1 / 61.7 & 51.2 / 60.3&-- / -- &-- / --& -- / --\\ \hline
     	 
%     	 MINDA (RC-base) & 58.3 / 66.3 & 52.3 / 63.0 & \bf 53.3 / \bf 62.7&56.1 / 63.7 &51.6 / 63.0& \bf 52.1 / \bf 61.7 \\
     	%MERDR (RC-base) & 56.9 / 64.9 & 52.3 / 63.0 & \bf 52.7 / \bf 62.2&54.7 / 62.3 &51.3 / 62.6& \bf 51.4 / \bf 61.0\\ 
     MuGER$^2$-base & 58.2 / 66.1 & 52.9 / 64.6 &  53.7 /  63.6&56.7 / 64.0 &52.3 / 63.9 &  52.8 /  62.5\\ 
    %  	 \ \ \ \ \  w/o constractive learning &56.6 / 64.2 &50.7 / 61.2 &51.7 / 60.8&-- / -- &-- / --& -- / -- \\

     %	  \ \ \ \ \    w/o DA & 38.6 / 47.0 &52.7 / 63.8& 45.9 / 55.7 &-- / -- &-- / --& -- / --\\
     	  %\ \ \ \ \  w/o iterative determination & 54.7 / 62.2 & 51.5 / 62.0 & 51.4 / 60.5 &-- / --&-- / -- &-- / --\\ 
     	 
     	% MINDA (RC-large)& 58.6 / 66.7 &54.1 / 65.2 &\bf 54.5 / \bf 64.2&56.4 / 64.1 &54.6 / 65.5& \bf 54.0 / \bf 63.3 \\
     %	MERDR (RC-large)& 57.4 / 65.3 &53.9 / 64.9 &\bf 53.9 / \bf 63.5&54.8 / 62.5&53.8 / 64.6 &\bf 52.8 / \bf 62.2\\
  %\textbf{MuGER$^2$-base (RC-Large)}& 58.9 / 66.6 &55.4 / 66.9 &\bf 55.4 / \bf 65.1&56.9 / 64.2 &55.9 / 66.8& \bf 54.8 / \bf 64.1\\
  %\textbf{MuGER$^2$-large (RC-base)}& 60.6 / 69.0 &53.2 / 65.5 &\bf 54.8 / \bf 65.3&58.0 / 66.0 &53.6 / 65.5& \textbf{54.0} / \textbf{64.1}\\ 
 MuGER$^2$-large & 60.9 / 69.2 &56.9 / 68.9 &\bf 57.1 / \bf 67.3&58.7 / 66.6 &57.1 / 68.6& \textbf{56.3} / \textbf{66.2}\\     	
     	% \ \ \ \ \    w/o MIN& 52.3 / 60.5 &51.5 / 62.5 & 50.5 / 60.2 &-- / -- &-- / --& -- / --\\
    %  	 \ \ \ \ \  w/o constractive learning & 57.2 / 64.8 &52.1 / 63.0 & 52.7 /  62.1&-- / -- &-- / --& -- / -- \\
     	 
     	% \ \ \ \ \       w/o DA & 42.0 / 50.3 &54.2 / 65.5 & 48.1 / 58.0 &-- / -- &-- / --& -- / --\\
    %  	 \ \ \ \ \  w/o iterative determination  & 55.5 / 62.7 &53.1 / 64.2& 52.6 / 62.0 &-- / -- &-- / --& -- / --\\ 
     	 \specialrule{0em}{1pt}{1pt}
     	 \bottomrule \bottomrule
	\end{tabular}
% 	}}
	\caption{The EM and F1 scores on HybridQA of different models.}
	\label{main}
\end{table*}

\begin{table}[ht]
\small
\setlength{\belowcaptionskip}{-0.3cm}
\centering
\begin{tabular}{l|cccc}
\toprule
\toprule
\specialrule{0em}{1pt}{1pt}
\textbf{Split}&\textbf{Train}&\textbf{Dev}&\textbf{Test}&\textbf{Total}\\
\specialrule{0em}{1pt}{1pt}
\midrule
\specialrule{0em}{1pt}{1pt}
In Passage&35,215&2,025&2,045 &39,285\\
In Table&26,803&1,349&1,346& 29,498\\
Compute&664&92&72&864 \\
Total&62,682&3,466&3,463&69,611 \\
\bottomrule
\bottomrule
\end{tabular}
\caption{Statistics of HybridQA dataset.}
\label{tab}
\end{table}
\section{Experiments}
\subsection{Implementation Details}
\subsubsection{Dataset}
To verify our proposed model, we conduct experiments on HybridQA\footnote{\url{https://github.com/wenhuchen/HybridQA}} \cite{chen-etal-2020-hybridqa}, a dataset of multi-hop question answering
over tabular and textual data.
% For each question, HybridQA provides a WiKiTable along with its hyperlinked Wikipedia passages as the knowledge.
% HybridQA provides a WiKiTable along with its hyperlinked Wikipedia passages as the knowledge to answer a question. 
The basic statistics of HybridQA are listed in Table~\ref{tab}. 
In the data split, `In-Table' means the answer is a table cell value, and `In-Passage' means the answer exists in a linked passage.
`Compute' means the answer is computed by performing numerical operations. In this paper, we mainly focus on the first two types.

\subsubsection{Settings} Two variants of
BERT \cite{devlin2018bert} are utilized in MuGER$^2$, namely MuGER$^2$-base and MuGER$^2$-large. The learning rate is set to $5$e-$5$ and the training batch size is set to $30$.
% In the unified evidence retriever, a training batch is subdivided into $5$ mini-batches for five types of evidence with a mini-batch size of $6$. 
% Each mini-batch contains $2$ same positive instances and $4$ sampled negative instances for contrastive learning.
A batch is divided into 4 groups  corresponding to the 4 kinds of evidence and each group contains 6 instances.

\subsection{Comparison with Previous Methods}
\subsubsection{Baselines}
\textbf{MQA-QG} \cite{pan2020unsupervised} is an unsupervised framework that generates multi-hop questions from the hybrid evidence and uses the generated questions to train the QA model.
\textbf{Table-Only} \cite{chen-etal-2020-hybridqa} only relies on the tabular information to find the answer by parsing the question into a symbolic form and executes it.
\textbf{Passage-Only} \cite{chen-etal-2020-hybridqa} only uses the hyperlinked passages to find the answer by retrieving related passages to perform the RC process.  
\textbf{Hybrider} \cite{chen-etal-2020-hybridqa} solves the HybridQA using a two-stage pipeline framework to retrieve a table cell and extract the answer in its value or linked passages.
\textbf{Dochopper} \cite{sun2021end}  builds a retrieval index for each cell and passage sentence, and perform RC on the retrieved evidence.

\subsubsection{Results and Analysis}
% We use the EM and F1 as the evaluation metrics to compare the question-answering performance of our MuGER$^2$ and the baseline models.
We use the EM and F1 as the evaluation metrics to compare the performance of our MuGER$^2$ and that of baselines.
As shown in Table~\ref{main}, MuGER$^2$-base outperforms strong baselines of HybridQA task and achieves significant improvements on both development and test sets.
Specifically, its performance exceeds the Hybrider-larger whose parameter scale is larger than it.
This indicates that our proposed multi-granularity evidence is beneficial for this task.
In addition, compared with the base model, MuGER$^2$-large further obtains gains.

% Both of them indicate that our proposed multi-granularity evidence is beneficial for this task. 
%In the ablation studies, we respectively show the effectiveness of each modules in our framework. All the ablation experiments are conducted on the develop set of HybridQA.

\subsection{Ablation Study}

\subsubsection{Stage-1: Retrieval}
\paragraph{Effect of Contrastive Learning}Without the contrastive learning (CL) within each granularity of evidence, the unified retriever is only trained  by the binary cross entropy objective in the joint way. 
The results in Table~\ref{cl} demonstrate that CL brings $0.8\%$ EM and $1.1\%$ F1 improvements to our model. These improvements are from the ability of CL to enhance the evidence representation learning by pushing away the positive  and negative evidence representation in the semantic space.

\paragraph{Effect of Joint Training}
In our method,  we adopt the joint training (JT) method to train the unified retriever to capture the relative information among different granularity. To prove the effectiveness of this design, we compare it with a straightforward way which trains four separate retrievers for different granularity evidence. For fairness, we turn off the contrastive learning  and only use the binary cross entropy objective to train the four separate retrievers and the unified retriever. The results in Table~\ref{cl} show that JT brings $1.6\%$ EM and $1.3\%$ F1  improvements to the HQA performance. 
%The reason is that JR allows the learning of different granularity evidence share the learned knowledge.  

Besides the end-to-end HQA performance, we further give the retrieval recall of the four different granularity evidence and the complete multi-granularity evidence in Figure~\ref{recall}.  The retrieval recall (R@1) indicates whether the retrieved top-$1$ candidate of the corresponding  evidence contains the answer. 
The results show that JT improves the retrieval performance on all these retrieval tasks. It is proved that sharing knowledge from different granularity brings benefit to the evidence retrieval. 
%\subsubsection{Retrieval Recall of Different Granularity Evidence} 

%Moreover, the multi-granularity evidence achieves the highest retrieval recall due to its higher answer coverage which achieves $0.98$. 
%For the mono-granularity evidence retrieval, the finer the granularity, the lower the recall.
%In addition, the retrieval recall of the cell(value) is significantly lower than the other types of evidence.
%This is because the cell(value) evidence usually has numeric values and has no linked passages, which makes it difficult to retrieve from huge candidates based on limited semantic information. 

%It is observed that the improvement is more significant on the in-table questions because such questions have less retrieval error tolerance than in-passage questions since sometimes the answer to in-passage questions are mentioned in multiple passages.

\begin{table}[t]
\setlength{\belowcaptionskip}{-0.4cm}
	\centering
	\small
	\linespread{1.8}
%	\resizebox{0.49 \textwidth}{!}{
    \setlength{\tabcolsep}{1mm}{
	\begin{tabular}{l|ccc}
		\toprule
		\toprule
		\specialrule{0em}{1pt}{1pt}
		\multirow{2}{*}{\bf Model} 

		                   &In-Table & In-Passage & Total   \\   
		                   \specialrule{0em}{1pt}{1pt}
		                   \cline{2-4} 
		                   \specialrule{0em}{1pt}{1pt}
		                   &EM / F1 & EM / F1 & EM / F1    \\
		                   \specialrule{0em}{1pt}{1pt}
		                   \midrule
		                   \specialrule{0em}{1pt}{1pt}
		 MuGER$^2$-base & 58.2 / 66.1 & 52.9 / 64.6 & \bf 53.7 / \bf 63.6 \\ 
		 \specialrule{0em}{1pt}{1pt}
	\midrule
\specialrule{0em}{1pt}{1pt}
		\ \ \ w/o CL &59.1 / 66.9 &50.6 / 62.2 &52.9 / 62.5 \\
 		 \ \ \ \ \ \ \ \ \ w/o JT & 54.2 / 61.9 &51.5 / 63.2& 51.3 / 61.2\\
% 	%	 \hline 
% %	\ \ \ \ \ \  w/o CD in DA & 53.7 / 61.6 & 53.5 / 64.3 & 52.2 / 61.6 \\
 \specialrule{0em}{1pt}{1pt}
	\midrule
\specialrule{0em}{1pt}{1pt}
     \ \ \ w/o E-SEL & 38.6 / 47.0 &52.7 / 63.8& 45.9 / 55.7\\
	%	 \ w/o ATR & 42.6 / 49.7 &46.9 / 57.3& 44.0 / 53.0\\
% 		  \specialrule{0em}{1pt}{1pt}
%      	  \hline
%      	  \specialrule{0em}{1pt}{1pt}
%		\bf MuGER$^2$ &  58.9 / 66.6 &55.4 / 66.9 &\bf 55.4 / \bf 65.1 \\
	%	\hline
	%	\ w/o CT & 59.6 / 66.7 &53.3 / 64.5 & 54.4 /  63.8 \\
	%	  \ w/o CT and JT& 54.6 / 62.2 &53.9 / 65.3 & 52.8 / 62.5\\
	%	 \hline
%		 \ \ \ \ \ \  w/o CD in DA & 54.2 / 62.1 & 55.4 / 66.6 & 53.5 / 63.2 \\

	%	 \ \ \ \ \ \       w/o DA & 42.0 / 50.3 &54.2 / 65.5 & 48.1 / 58.0\\
%    	\ w/o ATR & 51.2 / 58.3 &53.7 / 65.1 & 51.4 / 60.9\\
%		\ w/o ATR & 47.1 / 54.9 &51.3 / 62.2 & 48.4 / 57.8\\
		 
		\specialrule{0em}{1pt}{1pt}
		\bottomrule
		\bottomrule
        \specialrule{0em}{1pt}{1pt}
	\end{tabular}
	}
	\caption{Ablation results based on MuGER$^2$-base.}
	\label{cl}
\end{table}

\begin{figure}[t]
\setlength{\belowcaptionskip}{-0.3cm}
\centering
\includegraphics[width=0.48\textwidth]{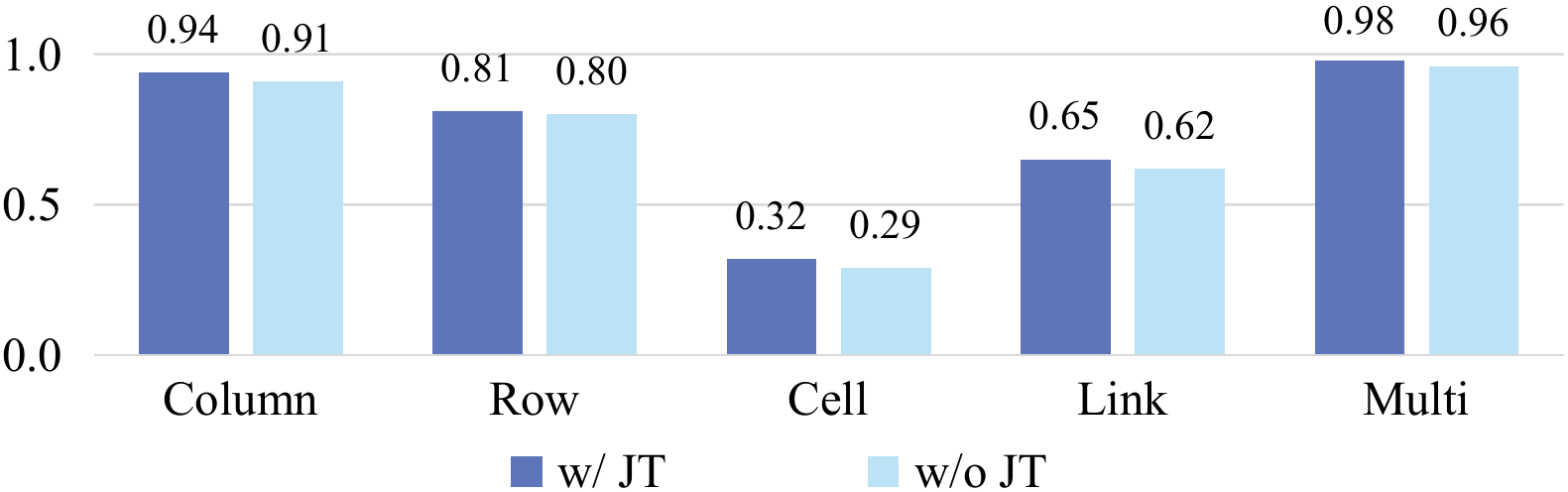}
% \caption{Retrieval precision (P) of each granularity of evidence in MuGER$^2$.}
\caption{R@1 of different granularity evidence.} 
\label{recall}  
\end{figure}

\subsubsection{Stage-2: Reasoning} 
\paragraph{Effect of Evidence Selector} To show the effectiveness of the evidence selector (E-SEL), we compare it with a straightforward answer reasoning method without any fine-grained evidence navigation. This method  simply flattens the retrieved top-1 multi-granularity evidence and feeding the sequence into the RC process. Results in Table~\ref{cl} show that the evidence selector increases the results by $7.8\%$ EM and $7.9\%$ F1 compared with the above reasoning method. This is due to its ability of accurately navigating the cell or link based on the multi-granularity evidence. The evidence selector reduces the burden of the RC model. In contrast, the above straightforward method brings redundant and noisy information which leads to weak RC performance.

\begin{table}[t]
\setlength{\belowcaptionskip}{-0.4cm}
	\centering
	\small
	\linespread{1.8}
	\begin{tabular}{l|ccc}
		\toprule
		\toprule
		\specialrule{0em}{1pt}{1pt}
		\multirow{2}{*}{\bf Granularity} 

		                   &In-Table & In-Passage & Total   \\   
		                   \specialrule{0em}{1pt}{1pt}
		                   \cline{2-4} 
		                   \specialrule{0em}{1pt}{1pt}
		                   &EM / F1 & EM / F1 & EM / F1    \\
		                   \specialrule{0em}{1pt}{1pt}
		                   \midrule
		                   \specialrule{0em}{1pt}{1pt}
	 %    Table & 12.8 / 18.2 & 11.6 / 15.4 &  12.4 /  16.8 \\ 
	 Column & 10.4 / 16.9  & 17.1 / 22.9  & 14.1 / 20.1 \\
 		Row &52.0 / 59.3 &49.6 / 60.1 & 49.3 / 58.3 \\
 		
 		Cell& 21.6 / 29.6 &0.0 / 1.4 & 8.4 / 12.3\\
        Link & 13.0 / 18.3 &40.6 / 50.1& 28.8 / 36.5\\
         Multi (ours) & 58.2 / 66.1 & 52.9 / 64.6 & \bf 53.7 / \bf 63.6 \\
% 		  \specialrule{0em}{1pt}{1pt}
%      	  \hline
%      	  \specialrule{0em}{1pt}{1pt}
%		     Table-g & 14.8 / 20.7 & 12.9 / 17.2 &  14.0 /  18.8 \\ 
%		Row-g &53.3 / 60.6 &52.4 / 62.9 &51.4 / 60.4 \\
%		Cell-g & 23.6 / 29.8 &41.4 / 51.4& 33.4 / 41.8\\
%		Link-g & 14.0 / 19.1 &41.7 / 51.8& 29.9 / 37.8\\
%		\bf MuGER$^2$ &  58.9 / 66.6 &55.4 / 66.9 &\bf 55.4 / \bf 65.1 \\ 
		\specialrule{0em}{1pt}{1pt}
		\bottomrule
		\bottomrule
		\specialrule{0em}{1pt}{1pt}
	\end{tabular}
	\caption{HQA results of different granularity evidence.}
	\label{evidence-g}
\end{table}
\subsection{Different Granularity Evidence\label{baselines}}
\subsubsection{Baselines}
To compare the HQA performance based on different granularity evidence, we conduct baselines for the four kinds of evidence including column, row, cell, and link. Specifically, the baselines of column, row and link granularity first retrieve the top-1 evidence of the selected granularity, and then flatten the retrieved evidence into a sequence to derive the answer by the RC model. While for the cell granularity, the value of the retrieved top-1 cell is directly returned as the answer. Note that, all the retrievers and the RC models of these baselines are initialized by the pre-trained BERT-base model.  

\begin{figure*}[t]
\setlength{\belowcaptionskip}{-0.4cm}
\centering
\includegraphics[width=0.99\textwidth]{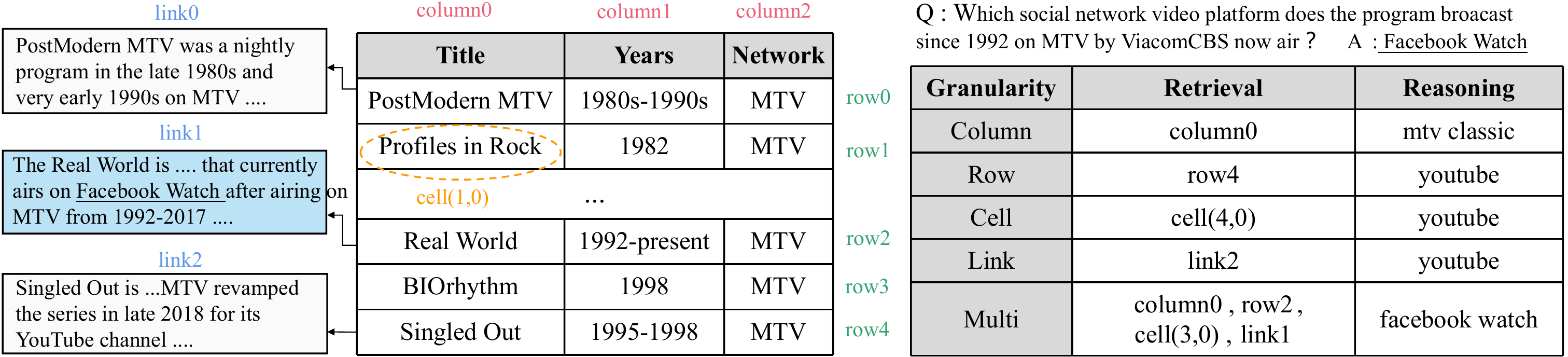}
\caption{An example of evidence retrieval and answer reasoning results based on  different granularity evidence.} 
\label{muger-case}  
\end{figure*}
\subsubsection{Results and Analysis}
The end-to-end HQA results based on different granularity evidence are shown in Table~\ref{evidence-g} and Figure~\ref{motivation}. 
The coarse-grained \textbf{Column} evidence achieves the best retrieval recall. This is because the question domains are easily mapped to a certain table header. However, reasoning the answer in a column is difficult because cells in a column are nearly indistinguishable in the absence of other semantic information, which leads to the lowest HQA performance.  
In contrast, the fine-grained \textbf{Cell} and \textbf{Link} evidence reduces the burden of answer reasoning but lowers the retrieval recall due to its huge candidates and limited semantic information. Moreover, only adopting the cell or link evidence cannot solve all questions which further causes low HQA performance.  
\textbf{Row} granularity achieves the best performance among these baselines since it balances the retrieval recall and RC performance to a certain extent. 
Compared with these baselines, our \textbf{Multi} granularity evidence brings remarkable improvements to the end-to-end HQA performance by preserving the evidence retrieval recall and the RC performance at the same time.

\section{Case Study}
Through statistics on the development set of HybridQA, we find there are 135 questions whose answers are only recalled by our multi-granularity evidence, but failed to be recalled by the coarse- and fine- grained evidence. 
An example of such a question is given in Figure~\ref{muger-case}.
The table along with linked passages on the left is the given knowledge to answer the question.
The answer \textit{facebook watch} is contained in \textit{link1} of \textit{cell(2,0)} where \textit{(2,0)} represents (\textit{row2, column0}).

 The table on the right lists the retrieved different granularity evidence. It is observed that the methods based on row, cell and link granularity evidence all retrieve incorrect evidence, including <\textit{row4, cell(4,0), link2}>. The reason is that the wrong evidence has high semantic similarity with the question which causes a higher ranking score. Although the column granularity based method retrieves the evidence <\textit{column0}> correctly, the reading comprehension model fails to derive the answer text accurately from the column.
 
In contrast, our method retrieves evidence including \textit{<column, row, cell, link>}. The evidence is correctly retrieved except for the \textit{cell} since the question is an in-passage question. The result demonstrates that the multi-granularity evidence improves the retrieval recall which benefits from learning the interactive information. The evidence selector further help to navigate the fine-grained cell or link based on the multi-granularity evidence. 
\section{Error Analysis}
\begin{figure}[t]
\setlength{\belowcaptionskip}{-0.4cm}
\centering
\includegraphics[width=0.45\textwidth]{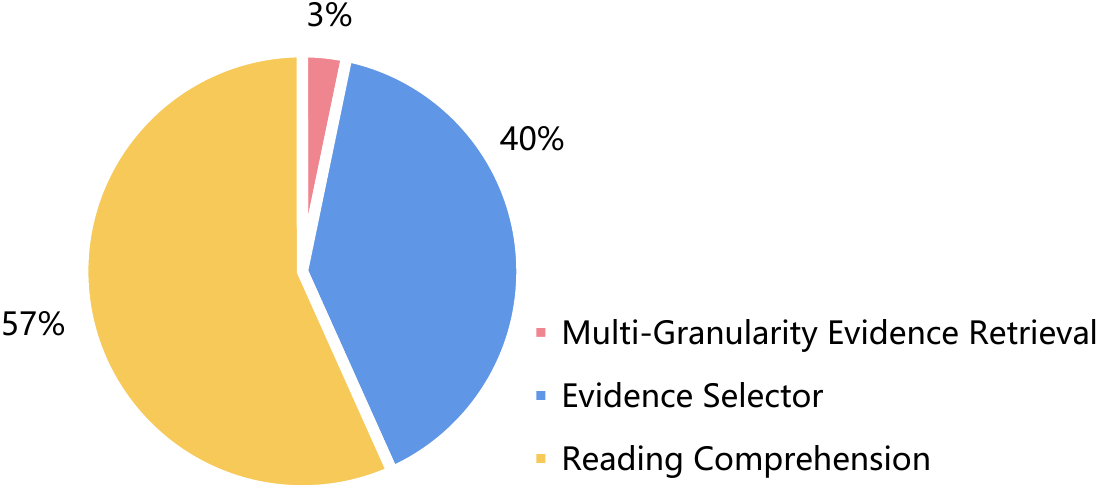}
\caption{Error of the three modules of MuGER$^2$-large.} 
\label{error}  
\vspace{-0.1cm}
\end{figure}
%\section{Appendix}
%\label{sec:appendix}
%\subsection{Error Analysis}

To further analyze our method, we also make statistics on error cases in the prediction of MuGER$^2$-large. The error statistics are based on the development set of HybridQA. 
As shown in Figure~\ref{error}, these cases can be divided into three categories according to the causes.
The first category is caused by evidence retrieval errors, which means the retrieved evidences overlook the answer.
According to our statistics, multi-granularity evidence has an advantage in improving coverage of answers.
Therefore, the proportion of this category is very low, only 3\%.
The second category is caused by the evidence selector, which accounts for about 40\% in the whole results. 
There are two reasons for the evidence selector to make the error. Firstly, although multi-granularity evidence improves the coverage, it also introduces noise.
As a result, it may also lead to a mis-navigation of the fine-grained cell or link.
In our results, it accounts for about 33\%.
In addition, for those questions which are offered correct cells or links, they also may be answered incorrectly due to the wrong judgment of the answer type.
It accounts for about 7\% in our results.
The last category is caused by the reading comprehension module.
Limited by the performance of the RC model, even if feed correct fine-grained evidence, the RC model also may produce the wrong answer.
According to our analysis, this category has the maximum proportion, accounting for about 57\% of the whole error cases.

\section{Discussion}
More recently, two novel approaches, MATE \cite{eisenschlos2021mate} and MITQA \cite{kumar2021multi}, are proposed and achieve significant results, respectively $70.0$ and $71.9$ F1 scores on the HybridQA test set. Specifically, MATE mainly focuses on strengthening the ability to encode large tables with multi-view attention. While MITQA designs a multi-instance training method based on distant supervision to filter the noise from multiple answer spans.
In contrast, our MuGER$^2$ puts more effort into integrating multi-granularity evidence for the HQA problem to enhance the reasoning ability. We think that the two novel approaches and our model pay attention to different perspectives for solving the HQA problems, and it’s probably to combine them to achieve better performance. %strengthens the transformer's ability to encode large tables with the multi-view attention, benefiting the tabular knowledge-based question-answering performance.

\section{Related Work}
Recently, many researchers attempt to tackle the question answering task using heterogeneous knowledge \cite{sun2019pullnet,sawant2019neural,zayats2021representations,li2021dual,xiong-etal-2019-improving,unirpg-zhou}. However, traditional researches only combine the information of heterogeneous knowledge, but fail to answer questions that require reasoning over heterogeneous data. To fill in the gap, the hybrid question answering (HQA) task together with the HybridQA  dataset is proposed by \cite{chen-etal-2020-hybridqa}. HybridQA provides a WiKiTable \cite{pasupat2015compositional} along with its hyperlinked Wikipedia passages \cite{rajpurkar-etal-2016-squad} as evidence for each question. 
%The HQA systems require to aggregate information from different forms to reason answers. 
Based on the HybridQA, \citet{chen2020open} further propose the OTT-QA dataset, which requires the system to retrieve relevant tables and text for the given questions.  \citet{zhu2021tat} and and \citet{chen-etal-2021-finqa}  propose TAT-QA and FinQA requiring numerical reasoning over the heterogeneous data. 

%This paper focus on the HybridQA task.

Existing works on HybridQA usually retrieve mono-granularity evidence from the heterogeneous data to derive the answer. Hybrider \cite{chen-etal-2020-hybridqa} proposes a two-phase pipeline framework to retrieve a table cell as the evidence and feeds its value and linked passages into an RC model to extract the answer.   Dochopper \cite{sun2021end} propose an end-to-end multi-hop retrieval method to directly retrieve a passage sentence or a cell value as the evidence. In addition, \citet{pan2020unsupervised} explore an unsupervised multi-hop QA model MQA-QG, which can generate human-like multi-hop training data from heterogeneous data resources. %Different from the previous work, this paper proposes the multi-granularity evidence retrieval and reasoning method to trade-off the answer recall and extraction accuracy.

\section{Conclusion}
We propose MuGER$^2$, a multi-granularity evidence retrieval and reasoning approach for hybrid question answering (HQA). In our method, a unified retriever is designed to learn multi-granularity evidence and an evidence selector is proposed to navigate fine-grained evidence for the reader.  Experiment results show that the MuGER$^2$ boosts the end-to-end HQA performance and outperforms the strong baselines  on the HybridQA benchmark.

\section*{Limitations}
In this paper, we focus on the hybrid question answering task, in which the answers to most questions can be extracted from the cell values or linked passages using the reading comprehension model. Although our MuGER$^2$ performs well on this task, one limitation is that it cannot answer the questions needing numerical operations, such as count, compare and etc. To enable our model to answer more complex questions, in future work we will develop the numerical reasoning capabilities of our model. 
\section*{Acknowledgment}
The authors would like to thank the anonymous
reviewers for their thoughtful comments.
The work of this paper is funded by the projects of National key research and development program of China (No.2020YFB1406902) and the National Key Research and Development Program of China (No. 2020AAA0108600). 
% Firstly, a unified retriever retrieves multi-granularity evidence with high answer recall from the table and linked passages. Afterwards, an answer type-aware reasoning process and an RC model are designed to derive the answer from the retrieved evidence.   Experiment results on the HybridQA dataset show that our method significantly   outperforms the strong baselines.
\bibliography{anthology,custom}
% \bibliography{MuGER2.new,custom}
\bibliographystyle{acl_natbib}
%\newpage
%\appendix

\end{document}